\documentclass[letterpaper]{article} 
\usepackage{aaai24}  
\usepackage{times}  
\usepackage{helvet}  
\usepackage{courier}  
\usepackage[hyphens]{url}  
\usepackage{graphicx} 
\usepackage{natbib}  
\usepackage{caption} 
\usepackage{algorithm}
\usepackage{algorithmic}
\usepackage{appendix}
\usepackage{newfloat}
\usepackage{listings}

\usepackage{microtype}
\usepackage{graphicx}
\usepackage{subfigure}
\usepackage{booktabs} 
\usepackage{amsmath} 

\DeclareCaptionStyle{ruled}{labelfont=normalfont,labelsep=colon,strut=off} 
\floatstyle{ruled}
\newfloat{listing}{tb}{lst}{}
\floatname{listing}{Listing}
\title{High Significant Fault Detection in Azure Core Workload Insights}
\author{
    Pranay Lohia,
    Laurent Bou\'e,
    Sharath Rangappa,
    Vijay Agneeswaran
}
\affiliations{
    Microsoft\\

    pranaylohia@microsoft.com,
    laurent.boue@microsoft.com,
    srangappa@microsoft.com,
    vagneeswaran@microsoft.com
}

\begin{document}

\maketitle

\begin{abstract}
Azure Core workload insights have time-series data with different metric units. Faults or Anomalies are observed in these time-series data owing to faults observed with respect to metric name, resources region, dimensions, and its dimension value associated with the data. For Azure Core, an important task is to highlight faults or anomalies to the user on a dashboard that they can perceive easily. The number of anomalies reported should be highly significant and in a limited number, e.g., 5-20 anomalies reported per hour. The reported anomalies will have significant user perception and high reconstruction error in any time-series forecasting model. Hence, our task is to automatically identify `high significant anomalies' and their associated information for user perception.
\end{abstract}

\section{Introduction}
\begin{figure*}[htb]
\begin{center}
\centerline{\includegraphics[width=\textwidth]{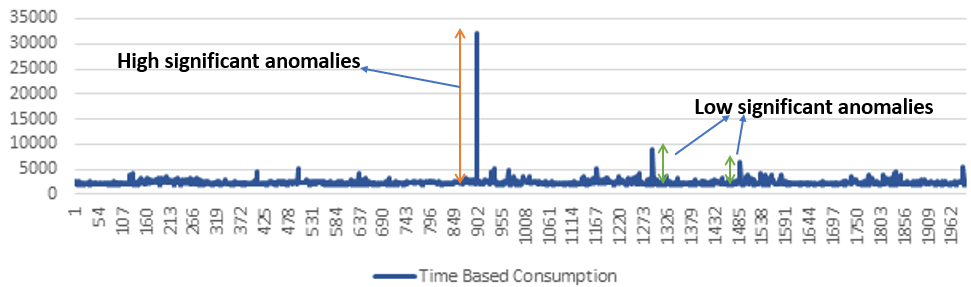}}
\caption{Time-series plot showcasing high and low significant anomalies}
\label{intro}
\end{center}
\end{figure*}

The idea of intelligent and workload-aware monitoring started in Azure. Based on the secondary research and customer discussions about the monitoring experience on Azure, we formulated our initial hypothesis: We believe more than 90\% Azure DevOps Engineers/IT Operations Teams are most overwhelmed \& frustrated about keeping the quality of service (QoS) of their applications (managing availability and maintainability of business solutions) on Azure because of:
\begin{itemize}
\item Proliferation of monitoring tools making analytical challenge
\item High volume of alerts leads to mismanagement and inability to pinpoint root causes. In ML parlance, this amounts to reducing the false positives
\item Reactive nature of the managing and monitoring solutions leading to customer downtime, and this also impacts customer experience
\end{itemize}
We have validated the hypothesis through 10 customer interviews, a customer survey (32 respondents), and Ignite conference (15+ customers). Extensive customer research helped us to solidify the hypothesis and unearthed pain points in the ‘As-Is’ journey of monitoring. We have identified three broader themes about the pain points: Issue Detection, Root Cause Analysis (RCA), and Early Mitigation Solutions. The current process of providing end-end (E2E) issue visibility to customers and addressing them is ‘Reactive’ \& ‘Slow’.  No solution is available to find the root cause at all levels and avoid issues.  The lack of an automated system to provide performance optimization recommendations led to multiple customer escalations due to slow RCA. We have proposed a solution to address the pain points across three broader themes of pain points. It is a one-stop place where customers can look for issues within their workload and recommendations to solve them. As a part of the grand solution, we are looking to start keeping customer workload as a central focus. The capability that we are contributing is that given a workload, we should be able to identify high significant outages/anomalies in the workload due to azure components/services. Figure~\ref{intro} outlines high and low significant anomalies in a sample time series from the azure core workload data. From red-outlined \textbf{high significant anomalies}, we can infer that they are very rare events in the data, easily perceived by users, and will have high reconstruction error in time-series forecasting models. Green outlined low significant anomalies are regularly occurring events, difficult to be perceived by users, and will have low reconstruction error in time-series forecasting models.

The current solution by the Azure Core team uses a manual filtering approach which yields a lot of anomalous/fault points (anomalies identified in order of thousands per hour) and consequent alerts.
The goal of the proposed approach is of two stages: Stage 1 involves internal private preview, while Stage 2 involves real customers.
\begin{itemize}
\item Stage 1: Yield anomalies with lesser false positives without reducing the ability to detect true positives, i.e., total anomalies detected should be less than 150 per hour.
\item Stage 2: Human-in-the-loop validation that only identifies high significant/human perceivable anomalies i.e., anomalies with high reconstruction error between actual and predicted. Hence, the total anomalies detected should be around 5-20 per hour.
\end{itemize}
Methods like WaveNet \cite{oord2016wavenet}, Temporal Fusion Transformers (TFT) \cite{lim2021temporal} and DeepAR \cite{salinas2020deepar} are able to achieve the goals of Stage 1. However, for Stage 2, we need to find highly significant anomalies and eliminate corner case anomalies where reconstruction loss is negligible. Prior methods have touched their `glass ceiling' configuration around 99.998\% quantile while methods like Extreme Value Theory (EVT) \cite{siffer2017anomaly} perform filtration in this high quantile range. By fine-tuning the hyper-parameter \textit{risk factor/quantile}, we are able to detect high significant anomalies.

Hence, our main contributions to the paper are listed as follows:
We deployed a system named MARIO and developed method to detect high significant and highly user-perceivable anomalies using Extreme Value Theory on top of time-series forecasting models with its application in detecting high significant faults in Azure Core workload insights data. We showcase the generalization of our method by obtaining efficient results on two bench-mark time-series datasets i.e., Electricity and Volatility dataset. We outperform two popular state-of-art methods i.e., TFT and DeepAR in identifying high significant anomalies in time-series data.

\section{Related Work}

Anomaly detection has been an important problem for which several machine-learning approaches have been tried in the last few years. The problem occurs in several contexts, including in the Internet of Things (IoT) where data from multiple devices keep flowing into edge devices or upstream cloud machines, where anomalies could mean failures of manufacturing devices or Internet nodes or in credit card fraud, or even cyber-security scenarios. It could occur in event/incident prediction, where it may be necessary to predict anomalies using log data. In the case of univariate data, the time series comprises a set of real-valued point-in-time observations, whereas multivariate data comprises a set of vectors, each of which comprises a series of point-in-time real-valued observations. The presence of noise, as opposed to real anomalies, complicates the problem and as a result, several techniques fail. Further, the ability of different methods to give a rank to the anomalies, based on the severity of the incidents, may be important to consumers of these algorithms.

Existing uni-variate algorithms are typically classified into a few categories, including the following:
\begin{itemize}
\item Statistical techniques or estimation techniques – The key techniques here include Median Absolute Deviation, B-splines, exponentially weighted moving averages, or even Gaussian Mixture Models.
\item Residual-based techniques – where the method analyzes the residuals obtained from different models to identify anomalies. Examples of these techniques include ARIMA models with exogenous inputs, STL decomposition, artificial neural networks, etc. 
\item Prediction-based techniques - like auto-regressive models and ARIMA models. Temporal prediction methods -\cite{gunnemann2014detecting} treats the base behavior as a latent multivariate auto-regressive process and overlays this with the sparse anomalous signal to generate the required observation. Density-based methods and distance/clustering techniques – as opposed to the above methods, which were based on defining anomalies based on an expected value, the density-based methods use some notion of distance and identifies anomalies as those values with the least neighbors. The notion of distance is not easy to quantify in temporal data, where attempts like \cite{angiulli2007detecting, angiulli2010distance} have been made. 
\end{itemize}
Most techniques above may work, but do not take into account the dependencies between the variables and may lead to a loss of information. Few techniques have been proposed to deal with this dependency. Dimensionality reduction techniques such as PCA – some approaches try out linear combinations of the input features and form a new set of uncorrelated variables and apply uni-variate techniques. Examples include \cite{papadimitriou2005streaming,baragona2007outliers}. Few techniques try to form a time-dependent single variable from the input variables, before applying uni-variate methods. One of the challenges Auto-Encoders face is that they are good at reconstructing the data without anomalies, but not so good when there are anomalies not encountered beforehand. USAD \cite{audibert2020usad} is a recent effort in this space that combines Auto-Encoders with Generative Adversarial Networks, which are used to discriminate whether the input data has anomalies, post which the data is passed to Auto-Encoders. Auto-Encoders address the stability problem of GANs too. Using the forecasting and pre-defined thresholds, it may be possible to determine anomalies in the data. Recent work in the forecasting space has been to introduce deep neural nets, which bring the advantage that the covariates across time series need not be manually specified but automatically inferred by the network to model complex group behavior. 

Further, by learning from similar items, efforts such as DeepAR \cite{salinas2020deepar} are able to provide forecasting for items for which no historical data is available. The history of deep neural networks for forecasting has been summarized in \cite{taieb2015bias}, where the authors investigate the best strategy for making multi-step predictions – up to possibly k steps, instead of just 1 step ahead. This is now been given a nomenclature as multi-horizon forecasting and the paper investigates the best strategy using RNNs and concludes that the direct strategy, which is to train a RNN with k-step target, instead of a single-step target is best. Another work in multi-horizon forecasting is \cite{wen2017multi} that presents a new ``forking-sequences'' training method which helps in learning across multiple series plus accommodates static features and temporal covariates as well as helps to model future events, such as the moving calendar (Chinese/Indian - lunar calendars). Another recent paper in this space is \cite{li2019enhancing}, where the authors investigate the weakness of applying a Transformer architecture to the forecasting problem and circumvent two common problems of Transformers, which are: first, point-wise dot product in canonical transformers are not sensitive to local context and second, the memory bottleneck of Transformers. They propose convolutional self-attention as a way of handling local sensitivity and LogSparse Transformer to handle memory bottlenecks. Another interesting recent work in this space is from Google, known as Temporal Fusion Transformers (TFTs) \cite{lim2021temporal}, which proposes an attention-based architecture for multi-horizon forecasting with novel interpretable insights into temporal dynamics. There are four important aspects of TFTs, which are 1. time-dependent LSTMs for processing the local context of the signal 2. interpretable multi-head attention for the long-term dependencies 3. Gating Residual Networks for efficient information flow with skip connections and gating layers. 4. variable selection, which can take inputs as static metadata, time-varying past inputs, a-priori known future inputs and outputs the static covariates (in encoded form) that need to be considered. Apart from DeepAR and TFT as state-of-art methods, there is a new hybrid method for predicting univariate and multivariate time series based on pattern forecasting \cite{castan2022new}. \cite{nie2022time} talks about using channel independence and sharing of weights across univariate series. Another work is using projection methods for functional time series forecasting \cite{elias2022projection}.


\section{System: Maintainability Availability Reliability Intelligence Ops (MARIO)}
MARIO is a deployed application that handles Azure Core workload insights. It is a service to empower the customer and the support team with product solutions to easily maintain/operate customer workloads using AIOps-powered Insights in Azure Core.

\section{System Details and Data Overview}

In this section, we present high-level system details, data collection, and data structure overview.
\subsection{Challenges and Opportunities with ML for Azure Core World}
\begin{itemize}
    \item Accelerate Time to Value: As teams across Microsoft leverage ML and AI to add value to their applications and offerings – a common issue is time to value is too long. While training and building the initial model may not take long (e.g., two weeks), operationalizing at scale with MLOps, governance, and tracking becomes a huge tax for individual teams. To solve this, we need economies of scale and realize value faster. We parse raw workload data from MARIO Service, create feature engineering, build a model to detect anomalies, and make it available as an offering for Azure MLaaS.
    \item Drive Adoption by Reducing Cognitive Load: How insights are surfaced in applications and experiences is a big factor in adoption and ultimately business impact. Providing too many faults/anomalies would lead to low usage and traction, as overwhelmed users do not have the time to figure out how to use these insights to improve their productivity. Users need the most important information surfaced to them to support their judgment, in line with their workflows. Also, whenever possible the action steps should be part of their workflow. In the case of Azure core workload, the system will highlight the anomaly when it occurred with confidence scores, upper and lower bounds for a series, and necessary attributes to help find the root cause and flag the anomaly as a false flag. This anomaly information along with poster action (correction/false flag) will be available to the reviewer for further awareness and review as needed. Finally, we periodically review both system-identified anomalies and actions taken on them. 
    \item Reduce System fault by Catching Anomalies Early: Identifying anomalies early with workload resource map application system faults could be proactively identified and appropriate actions can be taken/planned. 
    \item Increase Transparency through Confidence Score: Unless ML models have good transparency, users’ comfort and confidence with the model decreases causing slow/low adoption. They need transparent systems where the recommendations can be explained. Looking from an audit trail and review point of view how the model was used to mitigate risk needs to be very clear. To drive this high level of confidence we are adding confidence scores.
\end{itemize}


\subsection{Data Structure Overview}
The data that MARIO service collects from the customer tenant is from Azure Monitor insights. Data collected in Azure Monitor is stored in a time-series database that's optimized for analyzing timestamped data. Metrics are numerical values that describe some aspect of a system at a particular point in time. They are collected at regular intervals and are identified with a timestamp, a name, a value, and one or more defining labels. Metrics can be aggregated using a variety of algorithms, compared to other metrics, and analyzed for trends over time. Each set of metric values is a time series with the following: The time that the value was collected. The resource that the value is associated with. A name-space that acts like a category for the metric. A metric name. The value itself. Dimensions of a metric are name/value pairs that carry additional data to describe the metric value. For example, a metric called “Available disk space” might have a dimension called Drive with values C \& D. Dimension would allow viewing available disk space across all drives or for each drive individually.

Anomalies must be detected for each metric(M) for a given resource/dimension using time series of N-dimension values.
Azure monitor metrics are queried and posted to Mario storage endpoint.
JSON files have a common standard schema for all the resources. We have a pipeline to find the recent JSON files posted in Mario workspace and parse nested JSON data to a delta table(tabular) in CGAADLSDEV endpoint.
Table \ref{data summary} provides parsed data summary for a sample time frame highlighting different resource types, metrics, dimensions, dimension values, and a number of records associated with each meta-info.

\begin{table*}[t]
\vskip 0.15in
\begin{center}
\begin{small}
\begin{tabular}{lcccr}
\toprule
Unique Resource & Unique & Unique & Unique Dimension & Records \\
Type & Metrics & Dimensions & Values (Time-series) & \\
\midrule
Microsoft.DocumentDB & 18 & 13 & 1,302 & 13,914,350 \\
Microsoft.Storage & 7 & 6 & 168 & 4,215,037 \\
Microsoft.Network & 30 & 12 & 2,091 & 26,769,228 \\
Microsoft.ContainerService & 52 & 28 & 30,051 & 91,177,800 \\
Microsoft.ServiceBus & 19 & 2 & 33 & 3,000,996 \\
Microsoft.Sql & 26 & 1 & 2 & 406,443 \\
\midrule
Total & 152 & 62 & 33,647 & 139,483,854 \\
\bottomrule
\end{tabular}
\end{small}
\end{center}
\caption{Parsed data summary for sample time-frame}
\label{data summary}
\vskip -0.1in
\end{table*}

\section{Solution Overview}

For Stage 1, we are using Microsoft's Anomaly Detection as a Service (ADaaS) \cite{Krishna2022Probabilistic} as our baseline anomaly detection model. ADaaS uses Google’s WaveNet model \cite{oord2016wavenet} which is a dilated causal convolutional layer repurposed for forecasting and anomaly detection.


On top of Anomaly Detection as a Service (ADaaS)-uni-variate, we have done the following enhancements:
\begin{itemize}
\item Entity for Anomaly detection: will be a combination of columns for arriving at unique series of data with granularity: MSProviderIdentifier, MSProviderIdentifier1, ResouceRegion, MetricName, Dimension, DimensionValue, MetricValue, TimeStamp
\item We are handling data points at every 5-minute interval whereas current ADaaS does every day.
\item Training the model with 7day data for ‘Availability’ metrics with around 16k time-series. Inference data has 15k time series, and we are inferencing/validating our results every 1 hr.
\item We are employing the median imputation technique over the default zero imputation method used in ADaaS. This has given us a lesser Symmetric Mean Absolute Percentage Error (SMAPE) value. i.e., SMAPE with zero imputation is 29.22 whereas SMAPE with median imputation is 14.10.
\item We have increased the parameter of quantile filtering to 99.998\% with a z-score of 4.09. This has given us a lesser (149 anomalies which are 0.083\%) number of anomalies and helped us meet the requirements of Stage 1. \item Along with the anomaly’s detection, a confidence score is also provided with the anomalies based on the distance between the actual data values and Upper Confidence Limit/Lower Confidence Limit. The value range varies from 0 to 1 where 0 signifies least confidence and 1 highest confidence.
\end{itemize}
Table \ref{adaas_result} showcases the results of different sets of experiments conducted with Enhanced ADaaS. The last row highlights the best configuration. We are using an inference interval of 12 data points i.e., 1 hour, which makes the total data points for inference = 180216. 7 days training data has time-stamp range of \textit{`2022-05-15 00:00:00',`2022-05-22 00:00:00’}, while 7 days inference data has time-stamp range of \textit{`2022-05-22 00:00:00',`2022-05-29 00:00:00'}.

\begin{table*}[t]
\begin{center}
\begin{small}
\begin{tabular}{lcccccccr}
\toprule
Setup & Quantile\% & Time-series & Length & Training & SMAPE & Inference & \#Anomalies & Anomalies\%\\
& & count & of series & Time & & Time & & \\
\midrule
7 day & 99.99 & Train:16402 & 2016 & 45min & 29.22 & 16min & 174 & 0.1\\
zero impute & & Infer:15018 &&&&&&\\
\midrule
7 day & 99.99 & Train:16402 & 2016 & 45min & 14.10 & 16min & 164 & 0.09\\
median impute & & Infer:15018 &&&&&&\\
\midrule
7 day & 99.998 & Train:16402 & 2016 & 48min & 29.22 & 13min & 164 & 0.09\\
zero impute & & Infer:15018 &&&&&&\\
\midrule
7 day & 99.998 & Train:16402 & 2016 & 48min & 14.10  & 13min & 149 & 0.08\\
median impute & & Infer:15018 &&&&&&\\
\bottomrule
\end{tabular}
\end{small}
\end{center}
\caption{Experimental Results with Enhanced ADaaS}
\label{adaas_result}
\end{table*}

To increase our efficacy catering to Stage 2, we employ \textbf{Extreme Value Theory} (EVT) on top of Enhanced ADaaS. Extreme Value Theory \cite{siffer2017anomaly} gives a principled way to estimate the probability of rare events (most other frameworks only assign scores that lack the properties of mathematical probabilities). It can be applied to all signals regardless of underlying generating data distribution. It runs linearly in time/space; ideally suited for real-time streaming applications and/or limited hardware environments. Essentially, the idea is based on extreme value statistics which shows that the tail statistics of most distributions converge to universal distributions independent of the underlying generating process. Combining the \textit{Pickands–Balkema–De Haan theorem} \cite{wuthrich2004bivariate} with the Peak Over Threshold techniques demonstrates that the distribution can be parameterized by a \textit{Generalized Pareto Distribution – GPD} \cite{castillo1997fitting}. The y-axis and x-axis in the figure corresponds to nominal distribution value and data values respectively.

\begin{figure*}[htb]
\begin{center}
\centerline{\includegraphics{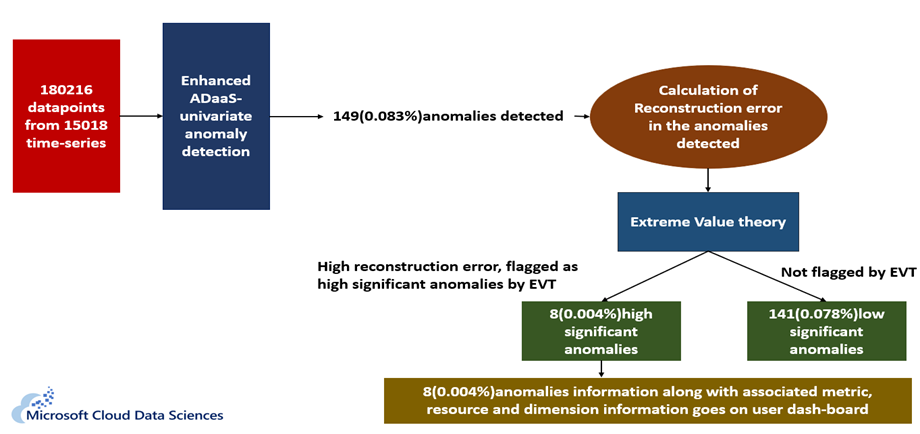}}
\caption{Enhanced ADaaS + EVT method flow diagram}
\label{flow}
\end{center}
\end{figure*}

Given a risk factor \begin{math}q\end{math}, a fixed quantile value \begin{math}t\end{math} (determined as the 98 percentiles in practice), \begin{math}n\end{math} as number of observations in a dataset, \begin{math}N_{t}\end{math} as the number of samples larger than \begin{math}t\end{math} and two parameters \begin{math} \gamma, \sigma\end{math} of the GPD distribution, the equation below:
\begin{equation}z_{q} \simeq t + \sigma/\gamma ((qn/N_{t})^{-\gamma} -1)
\end{equation}
gives us a threshold \begin{math}z_{q}\end{math} such that: \begin{math}P(X > z_{q}) < q\end{math}.
This inequality is now a strict mathematical statement about probabilities of rare events. Now the question arises: How to incorporate EVT as a high-pass filter in our algorithm? We observed that many anomalies have low reconstruction error i.e., they are close to the decision boundary. Hence, there is a further possibility to decrease the count of anomalies. We use EVT with \textbf{risk factor/quantile of (0.998)} filtering in the reconstruction loss function of Enhanced ADaaS and improve upon the results by decreasing the count of anomalies identified per hour.  We fine-tuned EVT's risk factor for experiments to be kept at 0.998 to be in sync with the human-in-loop validation discussed later. Beyond this quantile value, we were not observing any anomalies. Figure~\ref{flow} showcases \textbf{Enhanced ADaaS + EVT} method flow diagram where 180216 data points post Enhanced ADaaS produces 149 anomalies, which is further reduced to 8 high significant anomalies after employing EVT on top of Enhanced ADaaS.

\section{Result}
\begin{itemize}
\item We are getting 8 anomalies that have high reconstruction loss compared to 149 anomalies after employing EVT over enhanced ADaaS. These anomalies also have the highest confidence score.
\item Thus, we will be showcasing high priority anomalies or high significant anomalies (here, 8 anomalies) and the corresponding metric and resource information per hour for user perception on MARIO dashboard. 

\item The remaining set of 141 anomalies that have not been flagged by EVT will be showcased to users as low priority or low significant anomalies if demanded by them. These anomalies also have lower confidence scores compared to high significant anomalies.

\subsection{User-Study}
\item Human-in-the-loop validation or user-study feedback on the validity of the anomalies is taken on the service for subsequent improvements in our model. User-set include thousands of real customers of different Azure Workloads. These workloads include:
\begin{itemize}
    \item Device CARE: It is part of Supply Chain and their Cloud Native workload ‘Device Care’
     \item ACSS: Azure Centre for SAP solutions is an Azure offering that makes SAP a top-level workload on Azure
      \item DPS: Digital Professional Services create digital platforms that enables Industry Solutions, Support and Customer Success organizations to accelerate outcomes for commercial customers
\end{itemize}
We inferred the True Positive Rate to be 98\% for high significant anomalies. A low 4.92\% False Positive and 4.3\% False Negative Rate was observed for low significant anomalies, hence validating our solution.
\end{itemize}

\begin{table*}[t]
\begin{center}
\begin{small}
\begin{tabular}{lccccccr}
\toprule
Dataset & Method & Quantile\% & Anomaly Count & Anomaly\% & SMAPE & Precision & Recall\\
\midrule
& Enhanced ADaaS & 99.998 & 26 & 0.29 & 23.98 & 0.88 & 0.82\\
Electricity & Enhanced ADaaS & & & & & &\\
& +EVT & 99.998 & 9 & 0.10 & 23.26 & 0.92 & 0.94\\
\midrule
& Enhanced ADaaS & 99.998 & 21 & 4.12 & 45.57 & 0.84 & 0.85\\
Volatility & Enhanced ADaaS & & & & & &\\
& +EVT & 99.998 & 8 & 1.57 & 45.14 & 0.945 & 0.95\\
\bottomrule
\end{tabular}
\end{small}
\end{center}
\caption{Results on benchmark datasets}
\label{benchmark}
\end{table*}

\section{Generalisation of Method}
To showcase the generalization and applicability of our Enhanced ADaaS + EVT method, we experimented with two popular benchmark datasets:
\begin{itemize}
    \item \textbf{Electricity:} The UCI Electricity Load Diagrams Dataset, 
    \cite{misc_electricityloaddiagrams20112014_321} containing the electricity consumption of 370 customers – is aggregated on an hourly level. We use the past week (i.e., 168 hours) to find anomalies over the last 24 hours.
    \item \textbf{Volatility:} The OMI realized library \cite{heber2009oxford} contains daily realized volatility values of 31 stock indices computed from intraday data, along with their daily returns. We have used the same setup for our experiments as done in \cite{lim2021temporal}, We consider anomaly detection over the last week (i.e. 5 business days) using information over the past year (i.e. 252 business days).
\end{itemize}

In Table \ref{benchmark}, we have articulated anomalies and anomalies percentage identified from Electricity and Volatility datasets using both Enhanced ADaaS, and Enhanced ADaaS + EVT method. The latter i.e., Enhanced ADaaS + EVT method identifies very limited (0.1\% in Electricity and 1.57\% in Volatility dataset) anomalies, but with almost the same SMAPE value. SMAPE values noted in the table correspond to only data points identified as anomalous. It signifies that we are able to identify more of high significant anomalies which have high reconstruction error in the model and help in user perception. From Table, we can articulate that metrics like Precision and Recall have a high value in a range of more than 0.8 for Enhanced ADaaS method. This validates our proposition of having high true positives and less of false positive or negative cases. With Enhanced ADaaS + EVT method, we see an increment in these metrics to values higher than 0.92, with value of Recall even reaching 0.95 with volatility dataset.


\begin{table*}[t]
\begin{center}
\begin{small}
\begin{tabular}{lccccccr}
\toprule
Dataset & Method & Quantile\% & Anomaly Count & Anomaly\% & SMAPE & Precision & Recall\\
\midrule
& TFT & 99.998 & 197 & 2.22 & 23.49 & 0.533 & 0.445\\
Electricity & DeepAR & 99.998 & 250 & 2.82 & 23.84 & 0.567 & 0.434\\
& Enhanced ADaaS & & & & & &\\
& +EVT & 99.998 & \textbf{9} & \textbf{0.10} & \textbf{23.26} & \textbf{0.921} & \textbf{0.943}\\
\midrule
& TFT & 99.998 & 112 & 21.96 & 45.31 & 0.458 & 0.469\\
Volatility & DeepAR & 99.998 & 132 & 25.88 & 45.87 & 0.513 & 0.445\\
& Enhanced ADaaS & & & & & &\\
& +EVT & 99.998 & \textbf{8} & \textbf{1.57} & \textbf{45.14} & \textbf{0.945} & \textbf{0.952}\\
\bottomrule
\end{tabular}
\end{small}
\end{center}
\caption{Our method comparison results with state-of-art methods}
\label{compare}
\end{table*}

\begin{figure}[htb]
\begin{center}
\centerline{\includegraphics{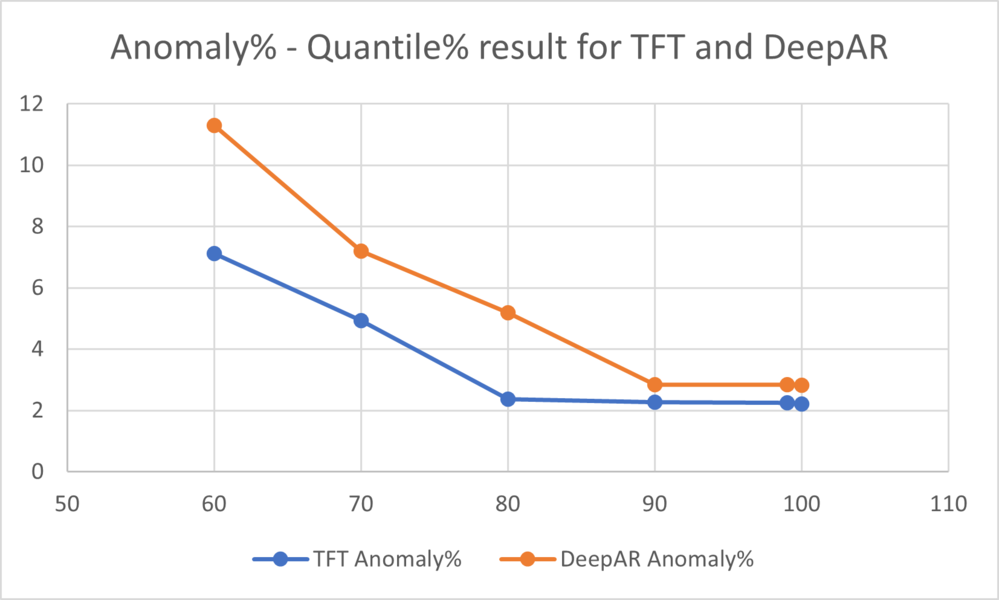}}
\caption{Plateaued performance plot of TFT and DeepAR with varying quantiles}
\label{prior_plot}
\end{center}
\end{figure}

\begin{figure}[htb]
\begin{center}
\centerline{\includegraphics{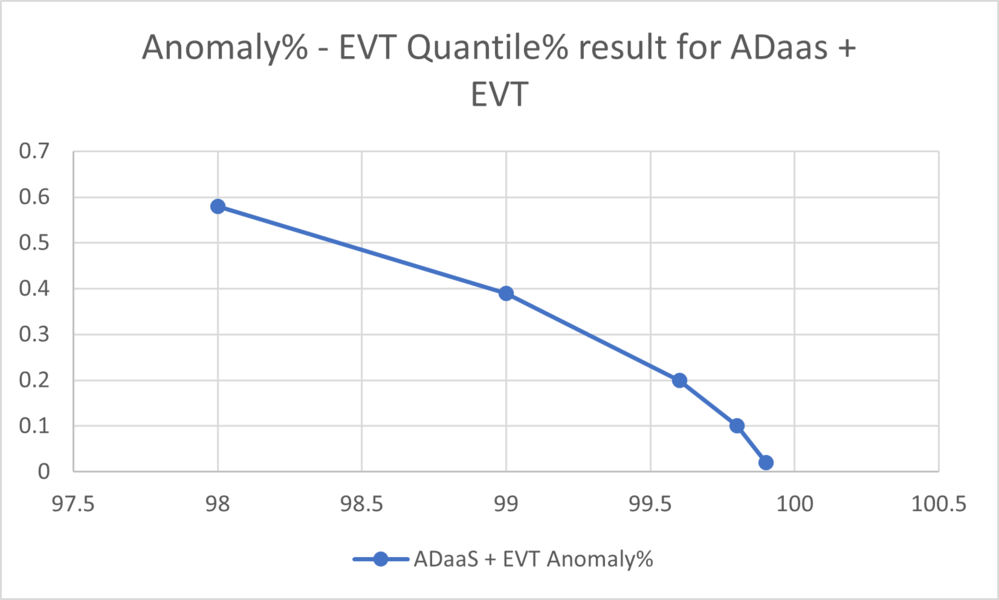}}
\caption{Growth performance plot of EVT with varying quantiles}
\label{evt_plot}
\end{center}
\end{figure}

\section{Comparison with State-of-Art}
To showcase better applicability of our method over state-of-art methods in finding high significant anomalies, we experimented and compared our results with Google's Temporal Fusion Transformers (TFT) \cite{lim2021temporal} and Amazon's DeepAR \cite{salinas2020deepar} on Electricity and Volatility Datasets. Before comparison, we experimented with TFT and DeepAR with a varying range of Quantile\% which is a hyper-parameter here. From Figure~\ref{prior_plot}, we observe that methods like TFT and DeepAR on Electricity Dataset are at their `glass ceiling' configuration around 99.998\% quantile value and have plateaued in their performance around high quantile range. Even with the increase in quantile\% (x-axis), we are not observing changes in anomalies\% (y-axis) observed. We have kept the same hyper-parameter i.e., 99.998\% quantile for our Enhanced ADaaS + EVT experiment setup. However, by using the benefit of EVT which actually starts filtration in these high quantile ranges, we have another hyper-parameter to fine-tune which is the risk factor/quantile to be used in EVT. In Figure~\ref{evt_plot}, we showcase that the high quantile range is the zone where EVT performs, and with an increase in EVT quantile\% (x-axis), Anomaly\% (y-axis) detected count decreases without plateauing.




From Table \ref{compare}, we can infer that methods like TFT and DeepAR identify more anomalies, i.e., they are finding less of high significant anomalies and more of less significant or corner cases anomalies that have low reconstruction error. However, our Enhanced ADaaS + EVT method identifies only 0.1\% in Electricity and 1.57\% in Volatility anomalies but with almost the same SMAPE values of 23.26 and 45.14 respectively, which signifies that our method is designed to identify precisely high significant anomalies which have high reconstruction error in the model and easier for user perception. Validation metrics like Precision and Recall are also higher in value in Enhanced ADaaS + EVT setup over TFT and DeepAR methods. We performed ablation study where we experimented with Enhanced ADaaS without EVT method. We got Precision-Recall numbers which were less than required threshold of 0.9 and several low-significant anomalies were observed. However, even in that case our method outperformed TFT and DeepAR. We experimented by using a parameter based DeepAR model which replaces the assumption of normal data distribution with a parameterized method which identifies the data distribution on the go. We got slight improvement (Precision: 0.62, Recall: 0.645 on Electricity Dataset) in results compared to DeepAR. Hence, even with ablations, our method is out-performing with competitors. As part of future work, we plan to come up with an algorithm that blends EVT with TFT and compare the performance with current work. We had also conducted our comparison study with Deep-Learning based MQRNN \cite{lim2021time} and DSSM \cite{rangapuram2018deep} also. Their metrics numbers on ‘Electricity dataset’ were as follows: MQRNN: Precision – 0.48, Recall – 0.40, DSSM: Precision - 0.493, Recall – 0.43. Since, TFT paper had already outperformed them, we choose to keep our study comprising two state-of-art models i.e., TFT and DeepAR.


\section{Conclusion}
Our solution provides automated fault/anomaly detection for Azure Core workload insights associated with different metrics and resources. We identify a limited set of highly significant anomalies with high reconstruction error which can be easily perceived on the user dashboard. Our solution with slight modifications is generic and can be applied to wider applications involving time-series anomaly or fault detection. There are some open-ended problems in terms of time and resource optimization which we aim to solve in the future. For products, we prioritized data with availability metrics. This data is stored on Azure Data Lake, and we are providing endpoint scoring for the training and inference process. Currently, we are training on 7 days of data and inferencing the last 1 hour. We have used Azure Databricks for parsing data and Azure ML Studio for building the models. In future: We plan to scale up the solution to incorporate all metrics of the workload insights data which amounts to 42k time series for 7 days of data. We plan to test our solution for the entire 3 months of data we have from the Azure Core workload. We are working towards a multi-variate anomaly detection-based solution in order to capture different relations between metrics and resources. Our system solution is in production and available as part of Microsoft Azure services.

\bibliography{CameraReady/LaTeX/main}

\begin{thebibliography}{23}
\providecommand{\natexlab}[1]{#1}

\bibitem[{Angiulli and Fassetti(2007)}]{angiulli2007detecting}
Angiulli, F.; and Fassetti, F. 2007.
\newblock Detecting distance-based outliers in streams of data.
\newblock In \emph{Proceedings of the sixteenth ACM conference on Conference on
  information and knowledge management}, 811--820.

\bibitem[{Angiulli and Fassetti(2010)}]{angiulli2010distance}
Angiulli, F.; and Fassetti, F. 2010.
\newblock Distance-based outlier queries in data streams: the novel task and
  algorithms.
\newblock \emph{Data Mining and Knowledge Discovery}, 20(2): 290--324.

\bibitem[{Audibert et~al.(2020)Audibert, Michiardi, Guyard, Marti, and
  Zuluaga}]{audibert2020usad}
Audibert, J.; Michiardi, P.; Guyard, F.; Marti, S.; and Zuluaga, M.~A. 2020.
\newblock Usad: Unsupervised anomaly detection on multivariate time series.
\newblock In \emph{Proceedings of the 26th ACM SIGKDD International Conference
  on Knowledge Discovery \& Data Mining}, 3395--3404.

\bibitem[{Baragona and Battaglia(2007)}]{baragona2007outliers}
Baragona, R.; and Battaglia, F. 2007.
\newblock Outliers detection in multivariate time series by independent
  component analysis.
\newblock \emph{Neural computation}, 19(7): 1962--1984.

\bibitem[{Cast{\'a}n-Lascorz et~al.(2022)Cast{\'a}n-Lascorz,
  Jim{\'e}nez-Herrera, Troncoso, and Asencio-Cort{\'e}s}]{castan2022new}
Cast{\'a}n-Lascorz, M.; Jim{\'e}nez-Herrera, P.; Troncoso, A.; and
  Asencio-Cort{\'e}s, G. 2022.
\newblock A new hybrid method for predicting univariate and multivariate time
  series based on pattern forecasting.
\newblock \emph{Information Sciences}, 586: 611--627.

\bibitem[{Castillo and Hadi(1997)}]{castillo1997fitting}
Castillo, E.; and Hadi, A.~S. 1997.
\newblock Fitting the generalized Pareto distribution to data.
\newblock \emph{Journal of the American Statistical Association}, 92(440):
  1609--1620.

\bibitem[{El{\'\i}as, Jim{\'e}nez, and Shang(2022)}]{elias2022projection}
El{\'\i}as, A.; Jim{\'e}nez, R.; and Shang, H.~L. 2022.
\newblock On projection methods for functional time series forecasting.
\newblock \emph{Journal of Multivariate Analysis}, 189: 104890.

\bibitem[{G{\"u}nnemann, G{\"u}nnemann, and
  Faloutsos(2014)}]{gunnemann2014detecting}
G{\"u}nnemann, S.; G{\"u}nnemann, N.; and Faloutsos, C. 2014.
\newblock Detecting anomalies in dynamic rating data: A robust probabilistic
  model for rating evolution.
\newblock In \emph{Proceedings of the 20th ACM SIGKDD international conference
  on Knowledge discovery and data mining}, 841--850.

\bibitem[{Heber et~al.(2009)Heber, Lunde, Shephard, and
  Sheppard}]{heber2009oxford}
Heber, G.; Lunde, A.; Shephard, N.; and Sheppard, K. 2009.
\newblock Oxford-Man Institute’s realized library.
\newblock \emph{Version 0.1, Oxford\&Man Institute, University of Oxford}.

\bibitem[{Krishna et~al.(2022)Krishna, Narayan, Khemka, Barrientos,
  Agneeswaran, and Kiran}]{Krishna2022Probabilistic}
Krishna, C.~S.; Narayan, S.; Khemka, S.; Barrientos, I.; Agneeswaran, V.; and
  Kiran, R. 2022.
\newblock Probabilistic Time-series Forecasting with Deep Autoregressive Flow
  Models.
\newblock In \emph{In Machine learning, AI and Data Science conference}. MSJAR.

\bibitem[{Li et~al.(2019)Li, Jin, Xuan, Zhou, Chen, Wang, and
  Yan}]{li2019enhancing}
Li, S.; Jin, X.; Xuan, Y.; Zhou, X.; Chen, W.; Wang, Y.-X.; and Yan, X. 2019.
\newblock Enhancing the locality and breaking the memory bottleneck of
  transformer on time series forecasting.
\newblock \emph{Advances in neural information processing systems}, 32.

\bibitem[{Lim et~al.(2021)Lim, Ar{\i}k, Loeff, and Pfister}]{lim2021temporal}
Lim, B.; Ar{\i}k, S.~{\"O}.; Loeff, N.; and Pfister, T. 2021.
\newblock Temporal fusion transformers for interpretable multi-horizon time
  series forecasting.
\newblock \emph{International Journal of Forecasting}, 37(4): 1748--1764.

\bibitem[{Lim and Zohren(2021)}]{lim2021time}
Lim, B.; and Zohren, S. 2021.
\newblock Time-series forecasting with deep learning: a survey.
\newblock \emph{Philosophical Transactions of the Royal Society A}, 379(2194):
  20200209.

\bibitem[{Nie et~al.(2022)Nie, Nguyen, Sinthong, and Kalagnanam}]{nie2022time}
Nie, Y.; Nguyen, N.~H.; Sinthong, P.; and Kalagnanam, J. 2022.
\newblock A time series is worth 64 words: Long-term forecasting with
  transformers.
\newblock \emph{arXiv preprint arXiv:2211.14730}.

\bibitem[{Oord et~al.(2016)Oord, Dieleman, Zen, Simonyan, Vinyals, Graves,
  Kalchbrenner, Senior, and Kavukcuoglu}]{oord2016wavenet}
Oord, A. v.~d.; Dieleman, S.; Zen, H.; Simonyan, K.; Vinyals, O.; Graves, A.;
  Kalchbrenner, N.; Senior, A.; and Kavukcuoglu, K. 2016.
\newblock Wavenet: A generative model for raw audio.
\newblock \emph{arXiv preprint arXiv:1609.03499}.

\bibitem[{Papadimitriou, Sun, and Faloutsos(2005)}]{papadimitriou2005streaming}
Papadimitriou, S.; Sun, J.; and Faloutsos, C. 2005.
\newblock Streaming pattern discovery in multiple time-series.
\newblock In \emph{Proceedings of the 31st international conference on Very
  large data bases}, 697--708. Citeseer.

\bibitem[{Rangapuram et~al.(2018)Rangapuram, Seeger, Gasthaus, Stella, Wang,
  and Januschowski}]{rangapuram2018deep}
Rangapuram, S.~S.; Seeger, M.~W.; Gasthaus, J.; Stella, L.; Wang, Y.; and
  Januschowski, T. 2018.
\newblock Deep state space models for time series forecasting.
\newblock \emph{Advances in neural information processing systems}, 31.

\bibitem[{Salinas et~al.(2020)Salinas, Flunkert, Gasthaus, and
  Januschowski}]{salinas2020deepar}
Salinas, D.; Flunkert, V.; Gasthaus, J.; and Januschowski, T. 2020.
\newblock DeepAR: Probabilistic forecasting with autoregressive recurrent
  networks.
\newblock \emph{International Journal of Forecasting}, 36(3): 1181--1191.

\bibitem[{Siffer et~al.(2017)Siffer, Fouque, Termier, and
  Largouet}]{siffer2017anomaly}
Siffer, A.; Fouque, P.-A.; Termier, A.; and Largouet, C. 2017.
\newblock Anomaly detection in streams with extreme value theory.
\newblock In \emph{Proceedings of the 23rd ACM SIGKDD International Conference
  on Knowledge Discovery and Data Mining}, 1067--1075.

\bibitem[{Taieb and Atiya(2015)}]{taieb2015bias}
Taieb, S.~B.; and Atiya, A.~F. 2015.
\newblock A bias and variance analysis for multistep-ahead time series
  forecasting.
\newblock \emph{IEEE transactions on neural networks and learning systems},
  27(1): 62--76.

\bibitem[{Trindade(2015)}]{misc_electricityloaddiagrams20112014_321}
Trindade, A. 2015.
\newblock {ElectricityLoadDiagrams20112014}.
\newblock UCI Machine Learning Repository.
\newblock {DOI}: https://doi.org/10.24432/C58C86.

\bibitem[{Wen et~al.(2017)Wen, Torkkola, Narayanaswamy, and
  Madeka}]{wen2017multi}
Wen, R.; Torkkola, K.; Narayanaswamy, B.; and Madeka, D. 2017.
\newblock A multi-horizon quantile recurrent forecaster.
\newblock \emph{arXiv preprint arXiv:1711.11053}.

\bibitem[{W{\"u}thrich(2004)}]{wuthrich2004bivariate}
W{\"u}thrich, M.~V. 2004.
\newblock Bivariate extension of the Pickands--Balkema--de Haan theorem.
\newblock In \emph{Annales de l'Institut Henri Poincare (B) Probability and
  Statistics}, volume 40, No. 1, 33--41. Elsevier.

\end{thebibliography}
\end{document}